\UseRawInputEncoding
\documentclass[letterpaper, 10 pt, journal, twoside]{IEEEtran} 
\title{Teleoperator-Aware and CBF-Based Adaptive Nonlinear MPC for Shared Autonomy in Obstacle Avoidance of Legged Robots}

\usepackage{amsmath,amssymb,amsfonts}
\usepackage{mathtools}
\usepackage{cite}
\usepackage{graphicx} % for pdf, bitmapped graphics files
\usepackage{epsfig} % for postscript graphics files
\usepackage{times} % assumes new font selection scheme installed
\usepackage{xcolor}
\usepackage{balance}
\usepackage{multicol}
\usepackage{array}
\usepackage{multirow}
\usepackage{booktabs}
\usepackage[colorlinks=true, allcolors=blue]{hyperref}

\newcommand{\Real}{\mathbb{R}}
\newcommand{\col}{\textrm{col}}
\newcommand{\Integer}{\mathbb{Z}_{\geq0}}

\newcommand{\diag}{\textrm{diag}}

\newcommand{\terminal}{\textrm{terminal}}
\newcommand{\stage}{\textrm{stage}}
\newcommand{\reff}{\mathrm{ref}}
\newcommand{\thr}{\textrm{th}}

\newcommand{\Xset}{\mathcal{X}}
\newcommand{\Uset}{\mathcal{U}}
\newcommand{\Oset}{\mathcal{O}}

\DeclareMathOperator*{\argmax}{arg\,max}
\DeclareMathOperator*{\argmin}{arg\,min}

\usepackage{etoolbox}

\makeatletter
\patchcmd{\@makecaption}{\scshape}{}{}{}
\makeatother

\AtBeginEnvironment{equation*}{\vspace{-3pt}}
\AtEndEnvironment{equation*}{\vspace{-3pt}}
\AtBeginEnvironment{equation}{\vspace{-3pt}}
\AtEndEnvironment{equation}{\vspace{-3pt}}
\AtBeginEnvironment{align}{\vspace{-3pt}}
\AtEndEnvironment{align}{\vspace{-3pt}}
\AtBeginEnvironment{alignat}{\vspace{-3pt}}
\AtEndEnvironment{alignat}{\vspace{-3pt}}
\newtheorem{example}{\textbf{Example}}
\newtheorem{remark}{\textbf{Remark}}
\newtheorem{problem}{\textbf{Problem}}

\author{Ruturaj Sambhus$^{1}$, Muneeb Ahmad$^{1}$, Basit Muhammad Imran$^{1}$, Sujith Vijayan$^{2}$, Dylan P. Losey$^{1}$, and Kaveh Akbari Hamed$^{1}$
\thanks{The work of R.~Sambhus, M.~Ahmad, S.~Vijayan, and K.~Akbari Hamed is partially supported by the NSF under Grant 2423725. The human-subject experiments were approved by the Virginia Tech Institutional Review Board under Protocol No. 24-1013. All participants provided informed consent.}
\thanks{$^{1}$R.~Sambhus, M.~Ahmad, B.~Imran, D.~Losey, and K.~Akbari Hamed (\textit{Corresponding Author}) are with the Department of Mechanical Engineering, Virginia Tech, Blacksburg, VA 24061, USA, {\tt\small \{ruturajsambhus, muneeb24, basit, losey, kavehakbarihamed\}@vt.edu}}% <-this % stops a space
\thanks{$^{2}$S.~Vijayan is with the School of Neuroscience, Virginia Tech, Blacksburg, VA 24061, USA, {\tt\small neuron99@vt.edu}}
}

\newcommand{\human}{\textrm{H}}
\newcommand{\robot}{\textrm{R}}

\begin{document}

\maketitle

%%%%%%%%%%%%%%%%%%%%%%%%%%%%%%%%%%%%%%%%%%%%%%%%%%%%%%%%%%%%%%%%%%%%%%%%%%%%%%%%

\begin{abstract}
Safe and effective collaboration between humans and autonomous legged robots is a fundamental challenge in shared autonomy, particularly for teleoperated systems navigating cluttered environments. Shared autonomy for legged robots remains challenging due to uncertain human commands, collision-avoidance requirements, and the need to coordinate high-level planning with dynamically feasible locomotion. This paper presents a teleoperator-aware, control barrier function (CBF)-based, adaptive nonlinear model predictive control (ANMPC) framework for shared autonomy of quadrupedal robots in obstacle-avoidance tasks. The framework employs a prescribed arbitration weight between human and robot actions while modeling human inputs using a noisily rational Boltzmann model, whose parameters are adapted online through a projected gradient descent (PGD) law based on observed joystick commands. CBF constraints are incorporated into a computationally efficient NMPC formulation to promote obstacle avoidance while accounting for predicted human actions. The hierarchical architecture integrates high-level CBF-based ANMPC (10 Hz), mid-level dynamics-aware NMPC (60 Hz) enforcing single rigid body (SRB) dynamics, and low-level nonlinear whole-body control (500 Hz). Human-in-the-loop hardware experiments with ten participants on a Unitree Go2 quadruped demonstrate real-time feasibility and collision-free navigation across 50 trials. Shared autonomy reduced average human control effort by 51\% and joystick activity from 88\% to 42\% of trial time, with comparable obstacle clearance and significantly higher perceived assistance than manual operation. Simulation-based ablation studies demonstrate improved obstacle clearance across operator rationality levels and up to a 66\% reduction in median human-action prediction error through online adaptation.
\end{abstract}

%Safety in HRI

%%%%%%%%%%%%%%%%%%%%%%%%%%%%%%%%%%%%%%%%%%%%%%%%%%%%%%%%%%%%%%%%%%%%%%%%%%%%%%%%

\vspace{-0.5em}
\section{Introduction}
\label{sec:Introduction}
\vspace{-0.3em}

Legged robots are increasingly envisioned for collaborative tasks in disaster response, search-and-rescue, and other safety-critical domains where full autonomy remains challenging. In such settings, \textit{shared autonomy} combines human teleoperation with autonomous decision-making.
However, uncertain human commands, complex locomotion dynamics, and the need to balance human control authority with obstacle avoidance pose significant challenges, particularly in cluttered environments.

%%%%%%%%%%%%%%%%%%%%%%%%%%%%%%%%%%%%%%%%%%%%%%%%%%%%%%%%%%%%%%%%%%%%%%%%%%%%%%%%

\vspace{-0.3em}
\subsection{Related Work}
\label{sec:Related_work}
\vspace{-0.3em}

Model predictive control (MPC) is widely used for trajectory optimization and control of legged robots, typically employing reduced-order template models \cite{Full_Koditschek_Template}, including the linear inverted pendulum (LIP) and its extensions \cite{kajita19991LIP,ALIP,SLIP,vLIP_Sreenath,HLIP_Ames}, as well as centroidal dynamics and single rigid body (SRB) models \cite{orin2013centroidal,Kim_Wensing_Convex_MPC_01,Wensing_VBL_HJB,Abhishek_Hae-Won_TRO,pandala2022robust,Leila_Hamed_RAL}. Although QP-based MPC with linearized models is computationally efficient, its limitations in capturing nonlinear dynamics and enforcing safety constraints have motivated nonlinear MPC (NMPC) frameworks for gait planning and safety-critical locomotion \cite{NMPC_Park_02,NMPC_CBF_Ames_Hutter,NMPC_CBF_Sreenath,Hutter_anymal_cbf_inWBC,Sleiman_RAL,BiConMP,Ruturaj_Locomanipulation_RAL,Basit_RAL}. 

Control barrier functions (CBFs) provide formal safety guarantees \cite{Ames_CBF} and are commonly implemented as zero-horizon QP-based controllers that minimally modify nominal control inputs while enforcing safety constraints \cite{CBF_MRS}. Integrating CBFs with MPC extends safety enforcement over a nonzero prediction horizon, enabling safe motion planning for quadrupedal robots \cite{Basit_RAL,NMPC_CBF_Sreenath,NMPC_CBF_Ames_Hutter} and car racing systems \cite{Koushil_CBFMPC}. However, most existing MPC/NMPC frameworks for legged
robots assume fully autonomous operation without explicitly incorporating human-in-the-loop shared autonomy or intent modeling.

%%%%%%%%%%%%%%%%%%%%%%%%%%%%%%%%%%%%%%%%%%%%%%%%%%%%%%%%%%%%%%%%%%%%%%%%%%%%%%%%

\begin{figure*}[t]
    \centering
    \includegraphics[width=0.95\linewidth]{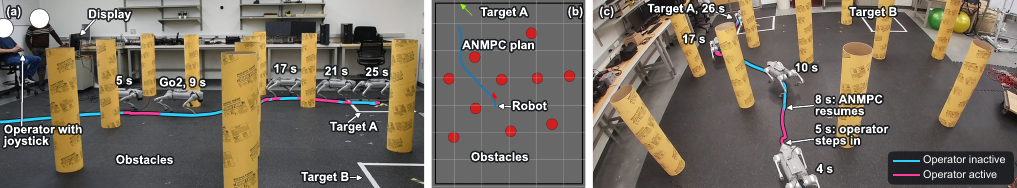}
    \vspace{-1.0em}
    \caption{Human-in-the-loop experiment using the proposed teleoperator-aware, CBF-based ANMPC. (a) Side-view snapshots of the Go2 navigating among 10 cylindrical obstacles. (b) GUI showing the robot, obstacles, target, and ANMPC plan. (c) Top-view snapshots highlighting operator intervention and inactivity. One of 50 trials involving 10 participants is shown. Videos are provided in the supplementary material.}
    \vspace{-1.5em}
    \label{fig:Snaphosts}
\end{figure*}

%%%%%%%%%%%%%%%%%%%%%%%%%%%%%%%%%%%%%%%%%%%%%%%%%%%%%%%%%%%%%%%%%%%%%%%%%%%%%%%%

Shared autonomy has been widely studied in the context of teleoperation, where the goal is to combine human intuition with robotic autonomy. In general, shared autonomy focuses on balancing the human's high-level knowledge (e.g., the goal the system should reach) with the robot's low-level automation (e.g., the motions to reach that goal) \cite{Dragan_Srinivasa_2013, Shervin_Shared_Autonomy_IJRR, hoffman2024inferring}. A variety of shared autonomy algorithms have been developed for robot arms \cite{jonnavittula2024sari, Muelling_2017}, quadrotors \cite{Reddy-RSS-18}, and smart vehicles \cite{Broad_2017}. However, these methods are not directly applicable to legged robots, where safe and effective operation requires solving highly dynamic, interconnected control problems in real time. For instance, if a human commands a quadruped to step laterally but its legs are not in a feasible stance, executing the command without accounting for the robot's dynamics could lead to instability or even falling.

Most relevant to our shared autonomy approach are methods combining intent blending with safety guarantees. \cite{SHARP_RA-L} introduced a stochastic optimal control framework coupling a noisily rational Boltzmann model of human behavior \cite{luce1959individual} with shielding-aware robust planning for safe human--robot interaction under uncertainty. This approach approximates the stochastic optimal control problem using scenario-tree-based stochastic MPC (SMPC). However, this framework and related methods primarily target autonomous driving and robotic manipulation, without addressing the challenges of legged locomotion, including high dimensionality and underactuation. \cite{Shared_control_quadruped_Rahul} developed an MPC framework incorporating human commands to control the front legs of a quadruped rescue robot while autonomously generating back-leg and center of mass (CoM) motions for static stability. However, it does not consider teleoperator-aware, CBF-based NMPC for obstacle avoidance. Additionally, \cite{Shared_control_humanoid_RAL} proposed a whole-body shared-control framework with bilateral force feedback for obstacle avoidance of a wheeled humanoid robot, but without CBF-based NMPC explicitly accounting for human intention.

%%%%%%%%%%%%%%%%%%%%%%%%%%%%%%%%%%%%%%%%%%%%%%%%%%%%%%%%%%%%%%%%%%%%%%%%%%%%%%%%

\vspace{-0.6em}
\subsection{Contributions}
\label{sec:Contributions}
\vspace{-0.3em}

The \textit{overarching goal} of this paper is to develop a computationally efficient, teleoperator-aware, CBF-based, and adaptive nonlinear model predictive control (ANMPC) framework for shared autonomy of quadrupedal robots performing obstacle-avoidance tasks. The main \textit{contributions} are as follows.

\textit{1. Teleoperator-aware CBF-based NMPC with implicit human action modeling:} Unlike prior shared autonomy and MPC approaches, the proposed framework introduces a novel formulation in which the human action is treated explicitly as an optimization variable within the NMPC problem and is constrained by the first-order optimality condition of a parameterized human value function derived from a noisily rational Boltzmann model. This enables seamless integration of human intent inference and CBF-based planning within a single optimization problem.

\textit{2. Online human-model adaptation and computationally efficient integration:}
The human value function combines quadratic goal-reaching terms with logarithmic safety barriers. Its unknown parameters are adapted online using a projected gradient descent (PGD) law derived through implicit differentiation of the human optimality condition. To address the computational challenges introduced by logarithmic barriers, a computationally efficient method is developed to integrate human-action prediction and CBF-based collision-avoidance constraints into the ANMPC
formulation.

\textit{3. Hierarchical control architecture for quadrupedal robots:} The overall control architecture is hierarchical. At the high level, the proposed CBF-based ANMPC generates blended human-robot velocity references at 10 Hz. At the mid level, a dynamics-aware NMPC enforces SRB dynamics to track these references at 60 Hz. At the low level, a nonlinear whole-body controller (WBC) imposes the full-order dynamics via quadratic programming to track the mid-level trajectories at 500 Hz (see Fig.~\ref{fig:Overview}). This layered architecture ensures consistent integration between high-level CBF-based planning and full-order robot dynamics execution.

\textit{4. Extensive hardware validation and ablation studies:}
The proposed framework is validated through real-time hardware experiments with ten participants on the Unitree Go2 quadruped (see Fig.~\ref{fig:Snaphosts}), demonstrating collision-free navigation across 50 trials, a 51\% reduction in average human control effort, and reduced joystick activity from 88\% to 42\% of trial time compared with manual operation. Additional simulation-based ablation studies evaluate the effects of operator rationality, CBF relaxation, and online human-model adaptation. The results demonstrate improved obstacle clearance across different operator rationality levels, comparable clearance under soft and hard CBFs, and a 66\% reduction in median human-action prediction error through online adaptation compared with a fixed-parameter model.

%%%%%%%%%%%%%%%%%%%%%%%%%%%%%%%%%%%%%%%%%%%%%%%%%%%%%%%%%%%%%%%%%%%%%%%%%%%%%%%%

\vspace{-0.5em}
\section{Problem Formulation}
\label{sec:problem_formulation}
\vspace{-0.3em}

\subsection{Robotic Agent and Environment}
\label{sec:robotic_agent_environment}
\vspace{-0.3em}

We consider a class of nonlinear, input-affine, discrete-time systems describing the robotic agent:
\begin{equation}\label{eq:robot_dyn}
    \!\!\! \Sigma \!:\! \left\{\begin{aligned}
        x(t+1) & = f(x(t),u(t)) = a(x(t)) + B(x(t))\,u(t)\\
        y(t)   & = C\,x(t),
    \end{aligned}
    \right.
\end{equation}
where $x\in\Xset\subset\Real^{n_x}$ and $u\in\Uset\subset\Real^{n_u}$ denote the state and control input, respectively, and $t\in\Integer:=\{0,1,\ldots\}$. The input-affine dynamics are characterized by $a(x)\in\Real^{n_x}$ and $B(x)\in\Real^{n_x\times n_u}$, with $f:\mathcal{X}\times\mathcal{U}\rightarrow\mathcal{X}$. The output $y=C\,x\in\Real^3$ represents the robot's planar position and yaw angle, where $C\in\Real^{3\times n_x}$.

%%%%%%%%%%%%%%%%%%%%%%%%%%%%%%%%%%%%%%%%%%%%%%%%%%%%%%%%%%%%%%%%%%%%%%%%%%%%%%%%

\begin{figure*}[t]
    \centering
    \includegraphics[width=\textwidth]{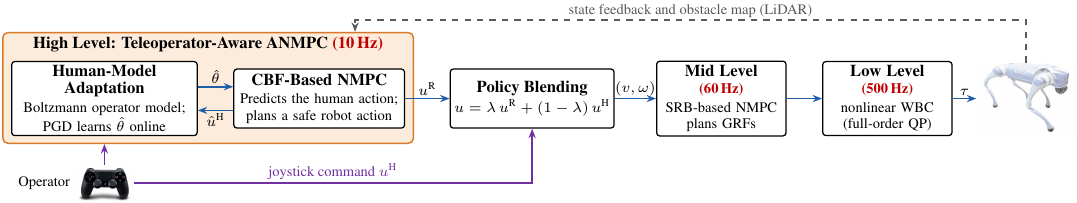}
    \vspace{-1.9em}
    \caption{Overview of the proposed hierarchical shared-autonomy framework. The high-level teleoperator-aware ANMPC predicts and adapts to the operator's joystick commands, and its robot action is blended with the human command ($\lambda$: robot authority).}
    \vspace{-1.3em}
    \label{fig:Overview}
\end{figure*}

%%%%%%%%%%%%%%%%%%%%%%%%%%%%%%%%%%%%%%%%%%%%%%%%%%%%%%%%%%%%%%%%%%%%%%%%%%%%%%%%

We consider shared autonomy in obstacle-avoidance tasks, where the control input is determined collaboratively by the human operator and the trajectory planning algorithm through linear policy blending. Specifically, the control input in \eqref{eq:robot_dyn} is given by
\begin{equation}\label{eq:weighted_sum}
u(t)=\lambda\, u^{\robot}(t)
+(1-\lambda)\,u^{\human}(t),
\end{equation}
where $u^{\human}(t)$ denotes the human operator's joystick commands, $u^{\robot}(t)$ is the control action generated by the proposed human-aware, CBF-based ANMPC algorithm, and $\lambda\in[0,1]$ determines the control authority. The cases $\lambda=0$ and $\lambda=1$ correspond to fully human and fully autonomous control, respectively.

\begin{example}[Kinematic Model]\label{example:kinematic_model}
Consider the unicycle model
\begin{equation}\label{eq:kinematic_car}
    \begin{bmatrix}
    \dot{p}_{x}\\
    \dot{p}_{y}\\
    \dot{\alpha}
    \end{bmatrix}
    =
    \begin{bmatrix}
    v\cos(\alpha)\\
    v\sin(\alpha)\\
    \omega
    \end{bmatrix},
\end{equation}
where $x:=\col(p_x,p_y,\alpha)\in\Real^3$ is the state vector, with $(p_x,p_y)$ denoting the planar position and $\alpha$ the yaw angle. The input $u:=\col(v,\omega)\in\Real^2$ consists of the linear velocity $v$ and angular velocity $\omega$, where $\col(\cdot)$ denotes the column operator. Applying forward Euler discretization yields the input-affine discrete-time model in \eqref{eq:robot_dyn}.
\end{example}

The target position of the robotic agent is assumed to be known to both the agent and the human operator, and is represented by the vector $g \in \Real^{3}$. This vector encodes the desired Cartesian coordinates of the target in the $xy$-plane together with the desired orientation (yaw angle). Static obstacles in the $xy$-plane are represented by their center points $o_{\ell} \in \Real^{2}$ for $\ell \in \Oset := \{1,\dots,n_{\mathcal{O}}\}$, where $n_{\Oset}$ denotes the total number of obstacles. 

%%%%%%%%%%%%%%%%%%%%%%%%%%%%%%%%%%%%%%%%%%%%%%%%%%%%%%%%%%%%%%%%%%%%%%%%%%%%%%%%

\vspace{-0.5em}
\subsection{Human Action Model}
\label{sec:human_action_model}
\vspace{-0.3em}

We model the teleoperator using a noisily rational Boltzmann model \cite{luce1959individual}, where actions are selected according to a parameterized value function $Q(x,u^{\human},\theta)$ with unknown parameters $\theta$ that are adapted online. The human action distribution is
\begin{equation}\label{eq:Boltzmann_model}
P\left(u^{\human}\mid x, \theta\right) =
\frac{e^{-\beta\,Q(x,u^{\human},\theta)}}
{\int e^{-\beta\,Q(x,\tilde{u}^{\human},\theta)}\, \textrm{d}\tilde{u}^{\human}} = \frac{e^{-\beta\,Q(x,u^{\human},\theta)}}{Z(x,\theta)},
\end{equation}
where $Q(x,u^{\human},\theta)$ denotes the human's state--action value function (equivalently, the negative reward), parameterized by a set of \textit{unknown} and time-varying parameters $\theta \in \Real^{n_{\theta}}$, $\beta>0$ is the rationality coefficient, and $Z(x,\theta):=\int e^{-\beta\,Q(x,\tilde{u}^{\human},\theta)}\, \textrm{d}\tilde{u}^{\human}$. Under this model, the teleoperator selects actions that are exponentially more likely to yield higher rewards (i.e., lower values of $Q$), that is,
\begin{alignat}{4}
    u^{\human\star}(x,\theta) := \argmax_{u^{\human}} P\left(u^{\human} \mid x,\theta\right) = \argmin_{u^{\human}} Q\left(x,u^{\human},\theta\right). \nonumber
\end{alignat}

%%%%%%%%%%%%%%%%%%%%%%%%%%%%%%%%%%%%%%%%%%%%%%%%%%%%%%%%%%%%%%%%%%%%%%%%%%%%%%%%

\subsubsection{Parametrization of the Human's Value Function}

We parameterize the human's state--action value function
using quadratic target-tracking and control-effort penalties,
together with logarithmic obstacle-avoidance terms.
Specifically,
\begin{alignat}{6}
    &Q\left(x,u^{\human},\theta\right) &&:=\,&&
    \left\|C\,f(x,u^{\human})-g\right\|_{M(\theta_1)}^2
    +\left\|u^{\human}\right\|_{M(\theta_2)}^2
    \nonumber\\
    & && &&-\theta_3\sum_{\ell=1}^{n_{\mathcal{O}}}
    \ln\frac{\left\|C_{xy}\,f(x,u^{\human})
    -o_{\ell}\right\|^2}{d_{\thr}^2},
    \label{eq:Q_func}
\end{alignat}
where $C\,f(x,u^{\human})\in\Real^3$ and $C_{xy}\,f(x,u^{\human})\in\Real^2$ denote the predicted robot pose and planar position under the human input, assuming no anticipation of robot assistance, and $d_{\thr}$ is the safety
distance threshold. Let $\theta:=\col(\theta_1,\theta_2,\theta_3)$, where $\theta_1,\theta_2$ are vectors and $\theta_3$ is a scalar, with $M(\theta_i):=\diag(\theta_i)$, $i=1,2$. Here, $\|z\|_M^2:=z^\top Mz$ and
$\|z\|^2:=z^\top z$. We require $M(\theta_1)\succ0$, $M(\theta_2)\succ0$,
and $\theta_3>0$.

%%%%%%%%%%%%%%%%%%%%%%%%%%%%%%%%%%%%%%%%%%%%%%%%%%%%%%%%%%%%%%%%%%%%%%%%%%%%%%%%

\subsubsection{Parameterized Human Action and Computational Challenge}

Given the parameterized value function in \eqref{eq:Q_func},
a stationary human action satisfies $\nabla_{u^{\human}}Q(x,u^{\human},\theta)=0$, yielding the algebraic constraint
\begin{alignat}{4}
    &\varphi\left(x,u^{\human},\theta\right)
    &&:=B^\top C^\top M(\theta_1)
    \left(C(a+B\,u^{\human})-g\right)
    \nonumber\\
    & &&+M(\theta_2)\,u^{\human}\nonumber\\
    & &&-\theta_3\sum_{\ell=1}^{n_{\mathcal{O}}}
    \frac{B^\top C_{xy}^\top
    \left(C_{xy}(a+B\,u^{\human})-o_{\ell}\right)}
    {\|C_{xy}(a+B\,u^{\human})-o_{\ell}\|^2}
    \nonumber\\
    & &&=0.\label{eq:algebraic_const}
\end{alignat}
For each $(x,\theta)$, the predicted human action $u^{\human\star}$ is obtained by solving $\varphi(x,u^{\human},\theta)=0$. Since this equation is nonlinear in $u^{\human}$, a closed-form solution is generally unavailable.
To maintain computational efficiency, we adopt a deterministic NMPC formulation rather than stochastic MPC (SMPC), without explicitly propagating uncertainty in the predicted human actions. As detailed in
Section~\ref{sec:HASC_ANMPC}, the proposed approach treats $u^{\human}$ as a decision variable subject to \eqref{eq:algebraic_const}, while adapting
$\theta$ online. The effect of operator rationality $\beta$ on stochastic human behavior is evaluated in the ablation studies in Section \ref{sec:Ablation_Study}.

\begin{remark}\label{remark:PD_Hessian}
Although $Q$ is not necessarily globally convex in $u^{\human}$, numerical evaluation confirmed that its Hessian $\nabla_{u^{\human}}^2 Q$ was positive definite at all computed stationary actions, using measured states and estimated parameters across all NMPC solves in Section~\ref{sec:Experiments}. The logarithmic terms reduced the smallest Hessian eigenvalue by at most 34\% relative to the quadratic terms alone. Thus, these actions are strict local minima of $Q$.
\end{remark}

%%%%%%%%%%%%%%%%%%%%%%%%%%%%%%%%%%%%%%%%%%%%%%%%%%%%%%%%%%%%%%%%%%%%%%%%%%%%%%%%

\vspace{-0.5em}
\subsection{Collision Safety and Problem Statement}
\label{sec:collision_safety_problem_statement}
\vspace{-0.3em}

To promote collision avoidance during human--robot collaboration, we incorporate discrete-time control barrier function (CBF)
constraints into the proposed ANMPC framework. For each obstacle $\ell\in\Oset$, we define the safety function
\begin{equation}\label{eq:hfun}
h_{\ell}(x):=\|C_{xy}\,x-o_{\ell}\|-d_{\thr},
\end{equation}
where $C_{xy}\,x\in\Real^2$ denotes the planar position of the robot, $o_{\ell}\in\Real^2$ is the center of obstacle $\ell$, and $d_{\thr}>0$ is the prescribed minimum separation distance. 

Following the discrete-time CBF formulation
\cite{DT-HOCBF}, the collision-avoidance constraints
are expressed as
\begin{equation}\label{eq:CBF_con}
h_{\ell}(x(t+1))-h_{\ell}(x(t))
\geq -\gamma\, h_{\ell}(x(t)),\quad \forall\ell\in\mathcal{O},
\end{equation}
where $\gamma\in(0,1]$ is the CBF parameter. The CBF condition in \eqref{eq:CBF_con}, together with the robot dynamics \eqref{eq:robot_dyn} and the weighted sum of the human and robot actions in
\eqref{eq:weighted_sum}, yields the following pointwise inequality constraint:
\begin{equation}\label{eq:CBF_ineq}
    \psi_{\ell}\left(x(t),\lambda\,u^{\robot}(t) + (1-\lambda)\,u^{\human}(t)\right) \geq0
\end{equation}
for all $t\in\Integer$ and $\ell\in\Oset$, where
$\psi_{\ell}(x,u):=h_{\ell}(f(x,u))-(1-\gamma)\,h_{\ell}(x)$.
For subsequent use, we stack the CBF functions for all obstacles and define $\psi:=\col(\psi_{\ell}:\ell\in\Oset)$.

\begin{remark}
For the kinematic model in Example~\ref{example:kinematic_model}, the one-step CBF constraint under forward Euler discretization depends on linear velocity but not explicitly on angular velocity. However, angular velocity affects the predicted heading and subsequent positions, allowing finite-horizon NMPC to account for both inputs in obstacle avoidance.
\end{remark}

\begin{problem}[Problem Statement]\label{problem:trajectory_planning}
We aim to design a computationally efficient, human-aware, CBF-based, and adaptive NMPC algorithm that computes the optimal robot action $u^{\robot}(t)\in\mathcal{U}^{\robot}\subset\Real^{n_{u}}$ to steer the robotic agent from an initial condition to the target point $g$, while (i) predicting the human action through the algebraic constraint \eqref{eq:algebraic_const}, (ii) incorporating relaxed CBF constraints to promote obstacle avoidance while accounting for uncertainty in the predicted human actions, and (iii) learning in real time the unknown parameters $\theta$ of the human state-action value function.
\end{problem}

%%%%%%%%%%%%%%%%%%%%%%%%%%%%%%%%%%%%%%%%%%%%%%%%%%%%%%%%%%%%%%%%%%%%%%%%%%%%%%%%

\vspace{-0.5em}
\section{Human-Aware, CBF-based ANMPC}
\label{sec:HASC_ANMPC}
\vspace{-0.3em}

This section presents the proposed human-aware, CBF-based ANMPC framework for addressing Problem~\ref{problem:trajectory_planning}. It consists of two loops: an \textit{inner loop} that optimizes robot actions using CBF-based NMPC with the current human-parameter estimate $\hat{\theta}(t)\in\Real^{n_{\theta}}$, and an \textit{outer loop} that updates this estimate online via projected gradient descent (PGD).

%%%%%%%%%%%%%%%%%%%%%%%%%%%%%%%%%%%%%%%%%%%%%%%%%%%%%%%%%%%%%%%%%%%%%%%%%%%%%%%%

\vspace{-0.5em}
\subsection{Inner Loop: CBF-based NMPC}
\label{sec:inner_loop}
\vspace{-0.3em}

The inner loop incorporates predicted human actions through the algebraic constraint \eqref{eq:algebraic_const} using the current parameter estimate $\hat{\theta}(t)$. Since the optimal human action $u^{\human\star}(x,\theta)$ generally lacks a closed-form solution, numerical approximations based on state and parameter discretization or machine learning could be
employed. However, substituting such approximations into the collaborative dynamics
\begin{equation}
x(t+1)=f\big(x(t),
\lambda\, u^{\robot}(t)+(1-\lambda)\,
u^{\human\star}(x(t),\hat{\theta}(t))\big)
\end{equation}
would introduce additional nonlinearities and increase the computational complexity of NMPC.

Instead, we formulate a computationally efficient CBF-based NMPC with relaxed safety constraints using penalized slack variables by incorporating the robot dynamics \eqref{eq:robot_dyn}, the blended control input \eqref{eq:weighted_sum}, and the algebraic constraint \eqref{eq:algebraic_const}. The decision variables include the predicted state, robot input, and human input trajectories, denoted by $x(\cdot)$, $u^{\robot}(\cdot)$, and $u^{\human}(\cdot)$, respectively, along with the CBF slack variables $\delta(\cdot)$. The human actions are determined implicitly by enforcing \eqref{eq:algebraic_const} as an equality constraint. Only the first optimized robot action is applied, while the predicted human input trajectory is used to satisfy the algebraic constraint and support parameter estimation. Given the current parameter
estimate $\hat{\theta}(t)$, the proposed NMPC problem is formulated as follows:
\begin{alignat}{6}
    & \min_{(x(\cdot),u^{\robot}(\cdot),u^{\human}(\cdot),\delta(\cdot))} && \hspace{-0.2cm} \mathcal{L}_{\terminal}^{\robot}(x_{t+N|t}) + \hspace{-0.1cm} \sum_{k=0}^{N-1} \mathcal{L}_{\stage}^{\robot}(x_{t+k|t},u^{\robot}_{t+k|t}) \nonumber\\
    & && + \sum_{k=0}^{N-1} \mathcal{L}_{\stage}^{\human}(u^{\human}_{t+k|t})  + \sum_{k=0}^{N-1} w\,\delta_{t+k|t}^{2}\nonumber\\
    & \textrm{s.t.} && \hspace{-2cm} x_{t+k+1|t} = f(x_{t+k|t},\lambda\,u^{\robot}_{t+k|t} + (1-\lambda)\,u^{\human}_{t+k|t}) \nonumber\\
    & && \hspace{-2cm} \varphi(x_{t+k|t},u^{\human}_{t+k|t},\hat{\theta}(t))=0  \hspace{1.8cm} \textrm{(Human Action)} \nonumber\\
    & && \hspace{-2cm} \psi(x_{t+k|t},\lambda\,u^{\robot}_{t+k|t} + (1-\lambda)\,u^{\human}_{t+k|t})\geq \delta_{t+k|t}\,\textbf{1} \hspace{0.3cm} \textrm{(CBF)} \nonumber\\
    & && \hspace{-2cm} \delta_{t+k|t} \leq 0, \,\, u^{\robot}_{t+k|t} \in \mathcal{U}^{\robot} \hspace{2.7cm} \textrm{(Feasibility)}\nonumber\\
    & && \hspace{-2cm} k=0,1,\cdots,N-1, \label{eq:CBF_based_NMPC}
\end{alignat}
where $N$ is the prediction horizon, and $x_{t+k|t}$, $u^{\robot}_{t+k|t}$,
$u^{\human}_{t+k|t}$, and $\delta_{t+k|t}$ denote the predicted states, robot inputs, human inputs, and CBF slack variables, respectively, with
$x_{t|t}=x(t)$. The terminal and stage costs are defined as
$\mathcal{L}_{\terminal}^{\robot}(x_{t+N|t})
:=\|x_{t+N|t}-x_{t+N|t}^{\reff}\|_{P^{\robot}}^2$,
$\mathcal{L}_{\stage}^{\robot}
(x_{t+k|t},u^{\robot}_{t+k|t})
:=\|x_{t+k|t}-x_{t+k|t}^{\reff}\|_{Q^{\robot}}^2
+\|u^{\robot}_{t+k|t}\|_{R^{\robot}}^2$,
and $\mathcal{L}_{\stage}^{\human}(u^{\human}_{t+k|t})
:=\|u^{\human}_{t+k|t}\|_{R^{\human}}^2$, where $x^{\reff}(\cdot)$ is the reference trajectory toward the goal, and $P^{\robot}$, $Q^{\robot}$,
$R^{\robot}$, and $R^{\human}$ are symmetric positive-definite weighting matrices. The scalar $w>0$ penalizes CBF slack variables, and
$\textbf{1}:=\col(1,\ldots,1)$ denotes the all-ones vector. The equality constraints enforce the collaborative dynamics \eqref{eq:robot_dyn} and
human-action stationarity \eqref{eq:algebraic_const}, while the inequality constraints impose relaxed CBF conditions \eqref{eq:CBF_con} and input
admissibility.

\begin{remark}
Strict CBF constraints may render ANMPC infeasible when human commands conflict with collision-avoidance requirements or control limits. We therefore introduce penalized slack variables to improve feasibility. Furthermore, discrepancies between actual and predicted human actions may violate constraints during execution, precluding unconditional collision-avoidance guarantees. Instead, safety is evaluated through simulations and hardware experiments, including a comparison of soft and hard CBF constraints in the ablation studies of Section~\ref{sec:Ablation_Study}.
\end{remark}

The CBF-based NMPC is solved in a receding-horizon fashion at 10 Hz for a given $\hat{\theta}(t)$. The first optimal robot action, defined as $\pi^{\robot}(x,\hat{\theta}(t)) := u^{\robot}_{t|t}$, is applied to the robotic system, while the estimated human action $u^{\human}_{t|t}$ is passed to the adaptive law in Section~\ref{sec:outer_loop} to update the parameter estimate $\hat{\theta}$ for the next time step.

%%%%%%%%%%%%%%%%%%%%%%%%%%%%%%%%%%%%%%%%%%%%%%%%%%%%%%%%%%%%%%%%%%%%%%%%%%%%%%%%

\vspace{-0.5em}
\subsection{Outer Loop: PGD-based Adaptive Law}
\label{sec:outer_loop}
\vspace{-0.3em}

The outer loop employs a PGD-based adaptive law to update the human value-function parameters using observed joystick commands $u^{\human}(t)$. At each time step, the parameter estimate is updated by minimizing the discrepancy between the predicted human action $u^{\human}_{t|t}$ and the measured action $u^{\human}(t)$ through the cost
\begin{equation}
J(\hat{\theta}):=
\tfrac{1}{2}
\left(u^{\human}_{t|t}-u^{\human}(t)\right)^\top
\Gamma
\left(u^{\human}_{t|t}-u^{\human}(t)\right),
\end{equation}
where $\Gamma=\Gamma^\top\succ0$ is a weighting matrix. The cost $J$ depends nonlinearly on $\hat{\theta}(t)$, since $u^{\human}_{t|t}$ is implicitly determined
by the parameterized algebraic constraint $\varphi(\cdot,\cdot,\hat{\theta}(t))=0$
in \eqref{eq:CBF_based_NMPC}.

Differentiating $J$ with respect to $\hat{\theta}$ yields
\begin{equation}\label{eq:dJ}
\frac{\partial J}{\partial\hat{\theta}}
=
\left(u^{\human}_{t|t}-u^{\human}(t)\right)^\top
\Gamma\frac{\partial u^{\human}_{t|t}}{\partial\hat{\theta}},
\end{equation}
where the measured human action $u^{\human}(t)$ is independent of $\hat{\theta}$. Since the NMPC enforces the algebraic constraint at time $t$,
\begin{equation}
\varphi(x(t),u^{\human}_{t|t},\hat{\theta}(t))=0,
\quad \forall\hat{\theta}(t)\in\Theta,
\end{equation}
where $\Theta\subset\Real^{n_{\theta}}$ denotes the feasible parameter set (e.g., nonnegative parameters). Implicit differentiation with respect to $\hat{\theta}$
gives
\begin{equation}\label{eq:diff_algebraic_const}
\frac{\partial\varphi}{\partial u^{\human}}
\frac{\partial u^{\human}_{t|t}}{\partial\hat{\theta}}
+
\frac{\partial\varphi}{\partial\hat{\theta}}=0.
\end{equation}
Since $\frac{\partial\varphi}{\partial u^{\human}}
=\frac{1}{2}\nabla_{u^{\human}}^2 Q$ is half the Hessian of the human value function $Q$, which was numerically verified to be positive definite at the computed stationary actions (Remark~\ref{remark:PD_Hessian}), substituting \eqref{eq:diff_algebraic_const} into \eqref{eq:dJ} yields
\begin{equation}
\frac{\partial J}{\partial\hat{\theta}}
=
-\left(u^{\human}_{t|t}-u^{\human}(t)\right)^\top
\Gamma
\left(\frac{\partial\varphi}{\partial u^{\human}}\right)^{-1}
\frac{\partial\varphi}{\partial\hat{\theta}}.
\end{equation}

The resulting PGD update is
\begin{alignat}{4}
&\eta(t+1)&&=\hat{\theta}(t)
-\mu(t)\nabla_{\hat{\theta}}J(\hat{\theta}(t))
\nonumber\\
& &&=\hat{\theta}(t)
+\mu(t)
\left(\frac{\partial\varphi}{\partial\hat{\theta}}\right)^\top
\left(\frac{\partial\varphi}{\partial u^{\human}}\right)^{-\top}
\Gamma
\left(u^{\human}_{t|t}-u^{\human}(t)\right)
\nonumber\\
&\hat{\theta}(t+1)&&=\Pi_{\Theta}(\eta(t+1))
=\argmin_{z\in\Theta}\|z-\eta(t+1)\|^2,
\label{eq:adaptive_law}
\end{alignat}
where $\mu(t)$ is the step size and $\Pi_{\Theta}(\cdot)$ denotes Euclidean projection onto $\Theta$. The updated estimate $\hat{\theta}(t+1)$ is used in the NMPC at the next time step.

%%%%%%%%%%%%%%%%%%%%%%%%%%%%%%%%%%%%%%%%%%%%%%%%%%%%%%%%%%%%%%%%%%%%%%%%%%%%%%%%

\vspace{-0.5em}
\subsection{Mid- and Low-Level Control}
\label{sec:Middle_and_Lower_Layers}
\vspace{-0.3em}

The mid- and low-level controllers are adapted from \cite{Basit_RAL,Ruturaj_Locomanipulation_RAL}. The high-level CBF-based ANMPC generates reference linear and angular velocities at 10 Hz by blending the optimized robot action with the measured human input:
\begin{equation}\label{eq:velocity_commands}
u(t) = \lambda\,\pi^{\robot}(x(t),\hat{\theta}(t))
+ (1-\lambda)\,u^{\human}(t).
\end{equation}
At the mid level, a dynamics-aware NMPC operating at 60 Hz tracks these references using the 6-DoF single rigid body (SRB) dynamics, optimizing ground reaction forces (GRFs) subject to dynamic and friction cone constraints \cite{Ruturaj_Locomanipulation_RAL}.
At the low level, a nonlinear WBC enforces full-order robot dynamics and tracks the prescribed SRB states and GRFs through a real-time QP operating at 500 Hz \cite{Randy_Paper_LCSS,pandala2022robust} (see Fig.~\ref{fig:Overview}).

%%%%%%%%%%%%%%%%%%%%%%%%%%%%%%%%%%%%%%%%%%%%%%%%%%%%%%%%%%%%%%%%%%%%%%%%%%%%%%%%

\vspace{-0.5em}
\section{Experiments}
\label{sec:Experiments}
\vspace{-0.2em}

\subsection{Setup and Controller Synthesis}
\label{sec:Setup}
\vspace{-0.2em}

We employ the Unitree Go2 quadruped for hardware validation and the RaiSim physics engine \cite{RAISIM} for numerical simulations. The Go2 weighs 15.0 kg, stands 0.28 m tall, and has 18 DoFs, including 12 actuated joints (hip pitch, hip roll, and knee pitch per leg). A Unitree L1 4D LiDAR
with $360^\circ\times90^\circ$ hemispherical coverage is used initially to map the environment and identify obstacles from deskewed point clouds. Human operators command the robot's torso linear and angular velocities using a Sony DualShock 4 controller. The proposed CBF-based ANMPC runs offboard in a multithreaded configuration on a workstation equipped with a 12th Gen Intel Core i9-12900F CPU and 64 GB RAM, communicating with the robot over a LAN.

\textit{CBF-based ANMPC Hyperparameters:}
The kinematic model \eqref{eq:kinematic_car} is discretized with $T_s=0.1$ s, using a prediction horizon of $N=100$ (equivalent to 10 s) and a blending factor of $\lambda=0.3$. The NMPC weights are $Q^{\robot}=\diag\{4,4,4\}$, $P^{\robot}=\diag\{40,40,40\}$, $R^{\robot}=\diag\{0.4,0.2\}$, $R^{\human}=\diag\{0.02,0.02\}$, and $w=10^3$. The CBF parameters are $\gamma=0.1$ and $d_{\thr}=0.5$ m.
For experimental validation, the human $Q$-function \eqref{eq:Q_func} uses four independent parameters: $\theta_1=\col(\theta_1^x,\theta_1^y,\theta_1^\alpha)$, with $\theta_1^x=\theta_1^y$ weighting planar target tracking and $\theta_1^\alpha$ weighting yaw tracking, where $M(\theta_1)=\diag(\theta_1)$. The scalar $\theta_2$ penalizes human linear and angular velocity commands with $M(\theta_2)=\theta_2 I_2$,
while $\theta_3$ weights the logarithmic obstacle-avoidance term. The initial estimates are $\hat{\theta}_1(0)=\col(0.4,0.4,5)$, $\hat{\theta}_2(0)=2$, and $\hat{\theta}_3(0)=2$. The PGD parameters in \eqref{eq:adaptive_law} are $\Gamma=\diag\{0.01,0.01\}$ and $\mu=1$. The blended commands \eqref{eq:velocity_commands} are saturated to $[-0.4,0.4]$ m/s for linear velocity and $[-0.8,0.8]$ rad/s for angular velocity. The mid-level NMPC and low-level nonlinear WBC hyperparameters follow \cite{Basit_RAL}. Consistent with Remark~\ref{remark:PD_Hessian}, the Hessian of $Q$ remained positive definite at the stationary human actions across all evaluated NMPC solves.

\textit{Real-Time Computation:}
The NLP in \eqref{eq:CBF_based_NMPC} involves $803$ decision variables, including $x(\cdot)$, $u^{\robot}(\cdot)$, $u^{\human}(\cdot)$, $\delta(\cdot)$, and the initial state constrained to measured feedback. It is implemented in CasADi \cite{CasADI} and solved using IPOPT \cite{IPOPT}, with a limit of $20$ iterations and warm-starting from the previous solution. The mean solve time was $17.4\pm3.5$~ms (mean $\pm$ standard deviation), with no infeasibility observed. The mid-level NMPC also uses CasADi, following \cite{Basit_RAL}.

%%%%%%%%%%%%%%%%%%%%%%%%%%%%%%%%%%%%%%%%%%%%%%%%%%%%%%%%%%%%%%%%%%%%%%%%%%%%%%%%

\vspace{-0.6em}
\subsection{Human-in-the-Loop Hardware Experiments}
\label{sec:Hardware_Experiments}
\vspace{-0.3em}

\textit{Experimental Setup:}
Human-in-the-loop experiments were conducted in an $8.1 \times 5.4$ m$^{2}$ laboratory space with 10 cylindrical obstacles (radius 0.125 m) and two target goals, $g_A$ and $g_B$ (see Fig.~\ref{fig:Snaphosts}). Each target specified the desired $xy$ coordinates and yaw angle of the robot and was marked on the floor with an arrow to guide participants. Ten subjects in the age group of 18-36, recruited from the university community with informed consent under an IRB-approved protocol, completed five trials after simulation practice. Two trials involved fully human control (baseline, $\lambda=0$) toward $g_A$ and $g_B$, and two used the proposed CBF-based ANMPC for shared autonomy toward the same targets. The cylindrical obstacles were detected using onboard LiDAR. An additional trial introduced ropes undetectable by LiDAR to evaluate human--robot collaboration under incomplete perception. 

During shared autonomy, the arbitration weight was set to $\lambda=0.3$ and temporarily reduced to $\lambda=0.1$ upon operator intervention (any joystick input outside a dead zone) to increase human control authority.  Operators stood beside the laboratory with an unobstructed view, while a screen displayed the robot's position and predicted trajectory in the $xy$-plane during shared autonomy, improving transparency and facilitating human--robot collaboration
(see Fig.~\ref{fig:Snaphosts}).

%%%%%%%%%%%%%%%%%%%%%%%%%%%%%%%%%%%%%%%%%%%%%%%%%%%%%%%%%%%%%%%%%%%%%%%%%%%%%%%%

\begin{figure}[t]
    \centering
    \includegraphics[width=0.85\linewidth]{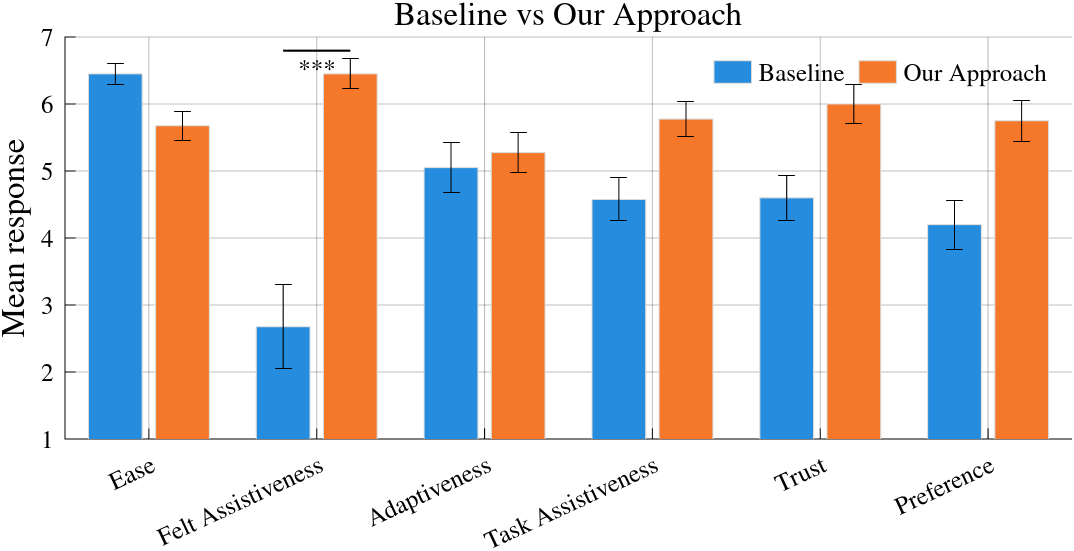}
    \vspace{-1.0em}
    \caption{Results of the Likert scale survey comparing the proposed shared control with the baseline control (fully human control).} 
    \vspace{-1.2em} % -2 in the original
    \label{fig:survey}
\end{figure}

%%%%%%%%%%%%%%%%%%%%%%%%%%%%%%%%%%%%%%%%%%%%%%%%%%%%%%%%%%%%%%%%%%%%%%%%%%%%%%%%

\textit{Quantitative Analysis:} 
Across all 50 trials, participants successfully navigated the quadrupedal robot toward the designated goals without collisions. After completing each experiment, participants filled out a survey questionnaire to evaluate the performance of each control method. Responses were recorded on a 7-point Likert scale ranging from Strongly Disagree (1) to Strongly Agree (7). Figure~\ref{fig:survey} summarizes the results of the post-experiment survey.

The survey results in Fig.~\ref{fig:survey} indicate that participants rated baseline control slightly higher in ease of use, potentially reflecting the greater direct control available during manual operation. In contrast, shared autonomy received substantially higher ratings for perceived assistance (paired $t$-test, $p<0.001$), as well as higher mean ratings for adaptiveness, task assistiveness, trust, and preference. The real-time GUI displaying the robot's state and predicted trajectory may have contributed to these favorable perceptions. Moreover, 9 of 10 participants preferred shared autonomy for prolonged operation (e.g., eight hours per day), suggesting its potential to reduce operator workload during extended missions.

\begin{figure}[t]
    \centering
    \includegraphics[width=0.85\linewidth]{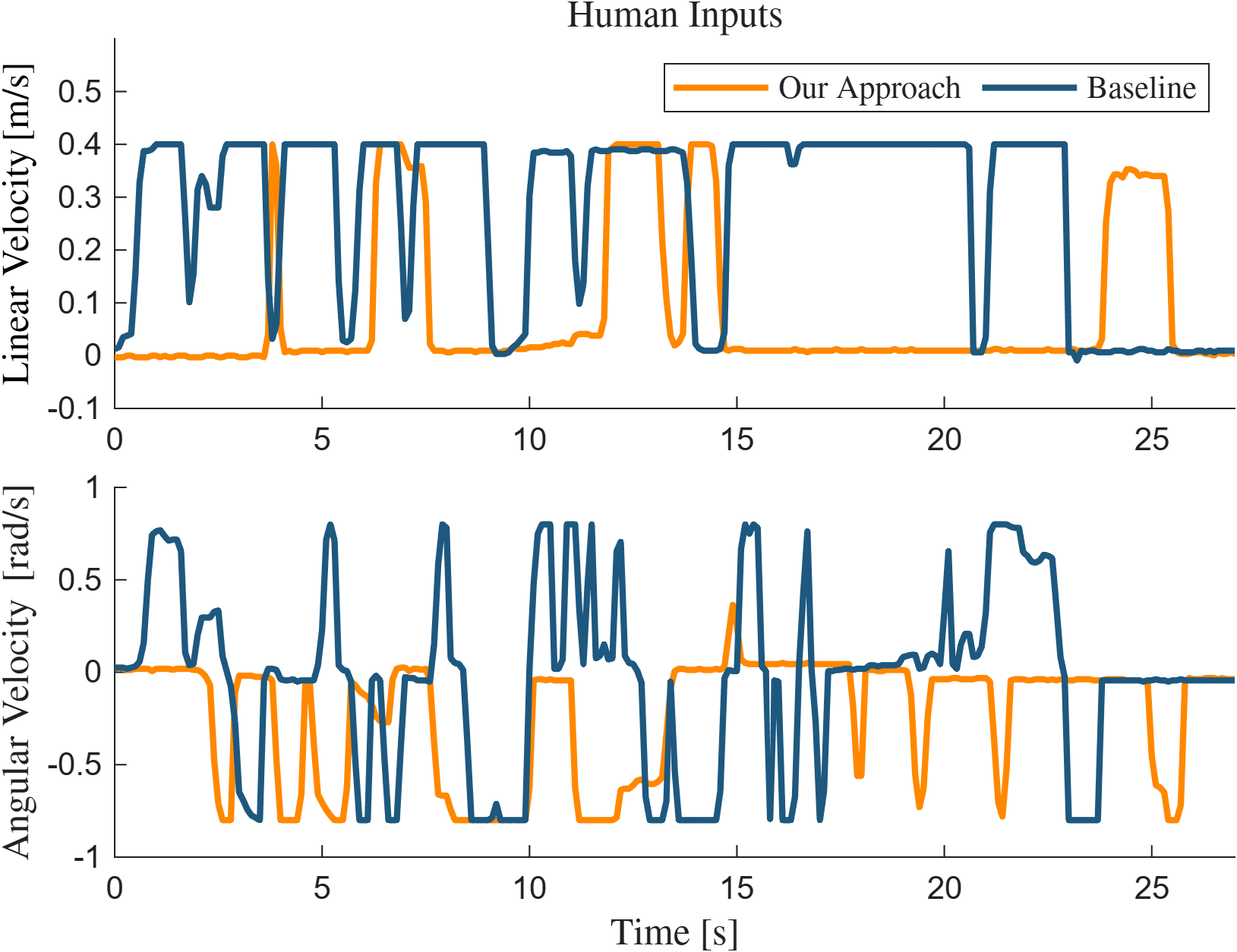}
    \vspace{-1.1em} % -1.3
    \caption{Plot of the measured human-commanded velocity inputs under the proposed shared autonomy framework and the baseline control method (fully human control), recorded from a representative subject.} 
    \vspace{-1.1em} % -0.8
    \label{fig:human_action}
\end{figure}

To quantitatively compare system responses under shared autonomy and baseline control, Fig.~\ref{fig:human_action} shows the human-commanded velocity inputs $u^{\human}=\col(v^{\human},\omega^{\human})$ from a representative subject. As observed, shared autonomy requires less human intervention. To quantify this across all trials, we compute the normalized human-effort metric
\begin{equation}
    J^{\human}:=\frac{1}{T_{\textrm{exp}}}\int_{0}^{T_{\textrm{exp}}}\|u^{\human}(t)\|^{2}\,\textrm{d}t,
\end{equation}
where $T_{\textrm{exp}}$ is the trial duration.  Averaged over all trials, $J^{\human}$ was $0.249$ for baseline control and $0.122$ for the proposed shared autonomy, indicating an approximately $51\%$ reduction in operator effort, while achieving similar completion times ($24.6\pm3.7$~s for the baseline versus $26.4\pm3.9$~s for the shared autonomy, $p=0.16$). Additionally, the average joystick activity, defined as the fraction of trial duration during which the joystick was deflected, decreased from $88\%$ under baseline control to $42\%$ under shared autonomy. This is consistent with the higher perceived assistance reported in the survey.

\begin{figure}[t]
    \centering
    \includegraphics[width=0.85\linewidth]{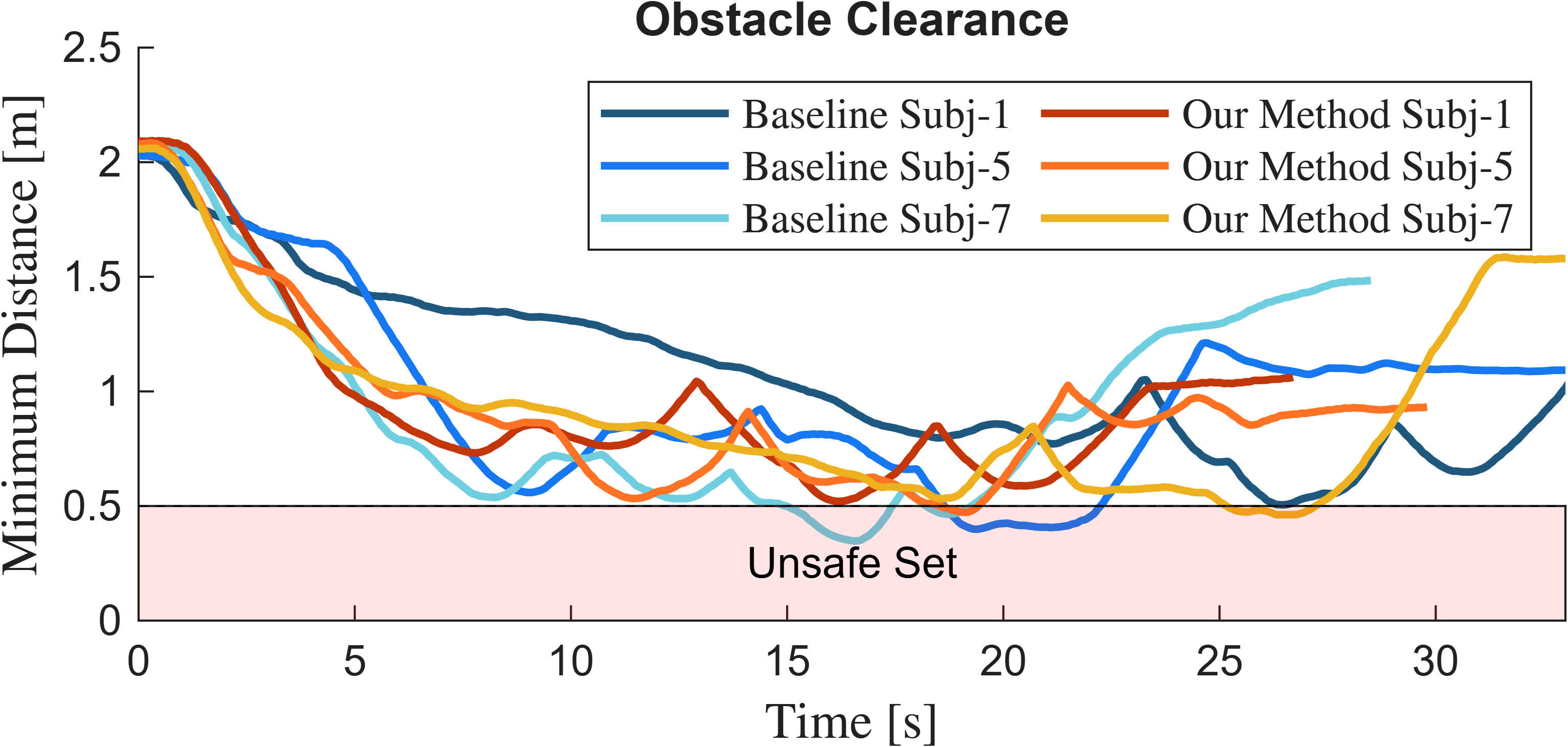}
    \vspace{-1.0em}
    \caption{Plot of the minimum distance between the robot and the nearest obstacle under the proposed shared autonomy and the baseline controller (fully human control) for three different subjects.} 
    \vspace{-1.5em} % -2
    \label{fig:distance}
\end{figure}

To evaluate safety performance, we analyze the minimum robot-obstacle distance across the baseline and the shared autonomy. Figure~\ref{fig:distance} shows the minimum robot-obstacle distance for three representative subjects. The smallest estimated distance to an obstacle center was 0.25~m with shared autonomy and 0.16~m with baseline control. A $1.7^\circ$ median heading offset between the state estimator and the LiDAR can shift an obstacle 5~m away by about 0.15~m. Experiment videos confirmed no physical contact in any trial.

Overall, however, neither the minimum distance ($0.44\pm0.09$~m versus $0.45\pm0.12$~m, $p=0.71$) nor the time spent inside $d_{\thr}$ ($1.8\pm1.6$~s versus $1.5\pm1.6$~s, $p=0.40$) differed significantly between shared autonomy and baseline control. Thus, shared autonomy reduced the operator effort without reducing obstacle clearance, but the user study does not show a statistically significant safety improvement. Since the operators' behavior cannot be controlled in a user study, Section~\ref{sec:Ablation_Study} further examines safety in simulation, where the operator rationality is varied systematically.

%%%%%%%%%%%%%%%%%%%%%%%%%%%%%%%%%%%%%%%%%%%%%%%%%%%%%%%%%%%%%%%%%%%%%%%%%%%%%%%%

%\vspace{1em}

%%%%%%%%%%%%%%%%%%%%%%%%%%%%%%%%%%%%%%%%%%%%%%%%%%%%%%%%%%%%%%%%%%%%%%%%%%%%%%%%

%%%%%%%%%%%%%%%%%%%%%%%%%%%%%%%%%%%%%%%%%%%%%%%%%%%%%%%%%%%%%%%%%%%%%%%%%%%%%%%%

\vspace{-0.5em}
\subsection{Ablation Study and Discussion}
\label{sec:Ablation_Study}
\vspace{-0.3em}

\textit{1) Effects of Operator Rationality and CBF Relaxation:}
To evaluate the proposed framework under different operator behaviors, we simulated human commands using the Boltzmann model~\eqref{eq:Boltzmann_model} in RaiSim with the Unitree Go2, ten obstacles, and the full control hierarchy. The operator samples commands from a discretized action space of 201 values per input in $[-0.5,0.5]$, using $\theta_1^x=\theta_1^y=8$, $\theta_1^\alpha=20$, $\theta_2=2$, and $\theta_3=1$. The rationality coefficient $\beta\in\{1,3,10,100,1000\}$ determines command variability, ranging from highly stochastic to nearly optimal actions. Since the human value function $Q$ in \eqref{eq:Q_func} evaluates only the next state, the simulated operator does not explicitly plan a collision-free route. We therefore select four intermediate waypoints along the route generated by the fully autonomous ANMPC ($\lambda=1$) toward $g=\col(6.3~\textrm{m},1.35~\textrm{m},\pi/4)$, switching within 0.3~m of each waypoint. This provides a common nominal route for the simulated operator and ANMPC, reducing route disagreement when comparing control strategies across different $\beta$. The ANMPC human model, however, uses only the final target, introducing a mismatch with the simulated operator. Future work will also investigate online waypoint generation and adaptation to better account for human navigation behavior.

The ANMPC parameters follow Section~\ref{sec:Setup}, except for $R^{\human}=\diag\{0.02,0.2\}$ and velocity limits of $\pm0.5$~m/s and $\pm0.5$~rad/s. The human model is initialized at $\hat{\theta}(0)=0.1\theta$ and adapted with $\Gamma=I_2$ and $\mu=1$. For each $\beta$, ten randomized trials were conducted under full human control ($\lambda=0$) and shared control ($\lambda=0.3$) with either soft or hard CBF constraints, where the latter removes the slack variables.

\begin{figure}[t]
    \centering
    \includegraphics[width=0.9\linewidth]{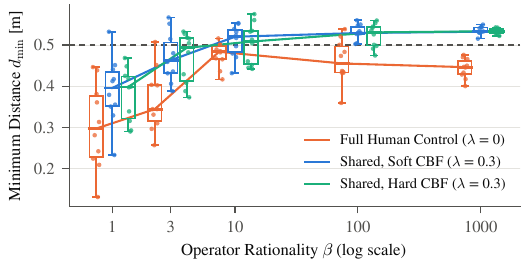}
    \vspace{-1.0em}
    \caption{Minimum obstacle distance $d_{\min}$ versus operator rationality $\beta$ under full human control ($\lambda=0$) and shared control ($\lambda=0.3$) with soft and hard CBFs, over ten trials per condition. Box plots show the distribution across trials; dashed line indicates the safety threshold $d_{\mathrm{th}}$.}
    \vspace{-1.3em} % -2
    \label{fig:sim_safety}
\end{figure}

Figure~\ref{fig:sim_safety} compares the minimum obstacle distance $d_{\min}$. Both shared-control formulations generally improved obstacle clearance over full human control, particularly for less rational operators. At $\beta=1$, only 5/10 trials reached the target within 130~s under full human control, compared with 10/10 trials under soft-CBF shared control. 
At $\beta=1000$, safety-threshold violations decreased from 10/10 (median $d_{\min}=0.45$~m) under full human control to 0/10 (median $d_{\min}=0.53$~m) under soft CBFs. 
Soft and hard CBFs yielded comparable clearance (median $d_{\min}$ differing by at most 0.03~m), completion times, and 100\% task completion under shared control. Hard CBFs caused more safety-margin violations at $\beta=100$ (3 vs.\ 0), but fewer violations at $\beta=3$ (5 vs.\ 7). Within the safety margin, the hard-CBF solver returned constraint-violating solutions in approximately 60\% of steps, comparable to the frequency of slack activation under soft CBFs. With a 20-iteration limit, the last iterate was applied regardless of constraint satisfaction, effectively allowing unpenalized
violations without improving clearance. Accordingly, we employ penalized slack variables in hardware experiments to improve optimization feasibility under conflicting control objectives, practical limitations, and uncertainty in human modeling.

%%%%%%%%%%%%%%%%%%%%%%%%%%%%%%%%%%%%%%%%%%%%%%%%%%%%%%%%%%%%%%%%%%%%%%%%%%%%%%%%

\begin{figure}[t]
    \centering
    \includegraphics[width=0.95\linewidth]{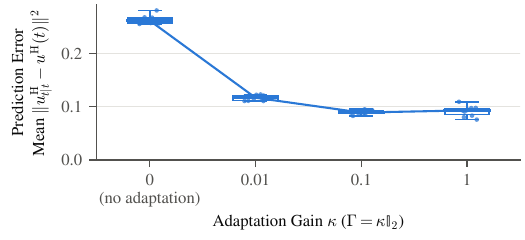}
    \vspace{-1.0em}
    \caption{Human-action prediction error versus adaptation gain $\kappa$ over ten trials per condition. Box plots show the distribution across trials, with $\kappa=0$ corresponding to no adaptation.}
    \vspace{-1.5em} % -2
    \label{fig:sim_adapt}
\end{figure}

\begin{figure}[t]
    \centering
    \includegraphics[width=0.9\linewidth]{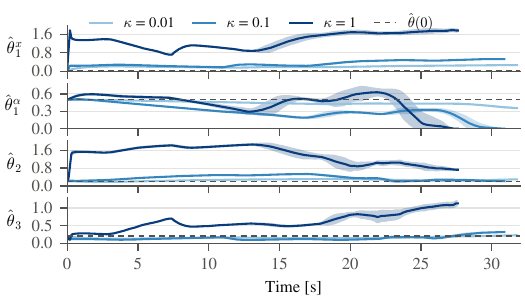}
    \vspace{-1.0em}
    \caption{Evolution of the estimated human-model parameters $\hat{\theta}$ for different adaptation gains $\kappa$. Solid lines show the mean over ten trials, shaded bands indicate one standard deviation, and dashed lines represent the initial parameter values.}
    \vspace{-1.5em} % -2
    \label{fig:sim_theta}
\end{figure}

%%%%%%%%%%%%%%%%%%%%%%%%%%%%%%%%%%%%%%%%%%%%%%%%%%%%%%%%%%%%%%%%%%%%%%%%%%%%%%%%

\textit{2) Effect of Online Human-Model Adaptation:}
To evaluate the effectiveness of online adaptation, we compared the proposed ANMPC with a fixed-parameter counterpart using the simulation setup described above, with $\beta=100$ and $\lambda=0.3$. The human model was initialized with $\hat{\theta}_1(0)=\col(0.04,0.04,0.5)$,
$\hat{\theta}_2(0)=\hat{\theta}_3(0)=0.2$, and adapted using~\eqref{eq:adaptive_law} with $\mu=1$ and $\Gamma=\kappa I_2$, where
$\kappa\in\{0,0.01,0.1,1\}$; $\kappa=0$ corresponds to no adaptation. Ten randomized trials were performed for each gain. The human-action prediction error was evaluated as
\begin{equation}
\bar{e}:=\frac{1}{T}\sum_{t=1}^{T}
\|u^{\human}_{t|t}-u^{\human}(t)\|^2,
\end{equation}
where $T$ denotes the number of control steps from the start of locomotion until the robot comes within 0.5~m of the target.

As shown in Fig.~\ref{fig:sim_adapt}, online adaptation substantially reduced the prediction error, with the median $\bar{e}$ decreasing from 0.262 without adaptation to 0.117, 0.089, and 0.093 for $\kappa=0.01$, 0.1, and 1, respectively. The lowest median prediction error was achieved at $\kappa=0.1$, corresponding to a 66\% reduction. Adaptation also reduced safety-margin violations from 8 of 10 trials without adaptation to 0, 2, and 0 of 10 trials for $\kappa=0.01, 0.1$, and 1, respectively. Figure~\ref{fig:sim_theta} illustrates the evolution of the estimated human-model parameters, which vary with the adaptation gain and do not necessarily converge to the operator's true weights. Since the human value function is linear in its parameters, the stationary human action is invariant under a common positive scaling of these parameters. Consequently, the adaptive law aims to improve human-action prediction rather than uniquely identify the underlying parameters. The observed parameter variations may also reflect the mismatch between the waypoint-following simulated operator and the human model, which considers only the final target. These results demonstrate that online adaptation improves human-action prediction and empirical safety performance despite model mismatch.

%%%%%%%%%%%%%%%%%%%%%%%%%%%%%%%%%%%%%%%%%%%%%%%%%%%%%%%%%%%%%%%%%%%%%%%%%%%%%%%%

\vspace{-0.5em}
\section{Conclusions}
\label{sec:Conclusions}
\vspace{-0.3em}

This paper presented a teleoperator-aware, CBF-based ANMPC framework for shared autonomy of quadrupedal robots performing obstacle avoidance. Human inputs are modeled using a noisily rational Boltzmann model, with parameters adapted online via a PGD law based on observed joystick commands. CBF-based collision-avoidance constraints are incorporated into a computationally efficient NMPC formulation to promote safe human--robot collaboration under uncertainty in human behavior. The hierarchical control architecture consists of a high-level CBF-based ANMPC (10 Hz), a mid-level NMPC enforcing reduced-order SRB dynamics (60 Hz), and a low-level nonlinear WBC (500 Hz) tracking full-order dynamics. Hardware experiments and a user study on the Unitree Go2 demonstrated real-time obstacle avoidance and a 51\% reduction in average human control effort compared with manual operation. Simulation-based ablation studies further demonstrated improved obstacle clearance across different operator rationality levels and up to a 66\% reduction in median human-action prediction error through online adaptation compared with a fixed-parameter model. Future work will investigate adaptive arbitration strategies in which the blending coefficient $\lambda$ is updated online based on safety conditions to further improve human--robot collaboration.

%%%%%%%%%%%%%%%%%%%%%%%%%%%%%%%%%%%%%%%%%%%%%%%%%%%%%%%%%%%%%%%%%%%%%%%%%%%%%%%%


%%%%%%%%%%%%%%%%%%%%%%%%%%%%%%%%%%%%%%%%%%%%%%%%%%%%%%%%%%%%%%%%%%%%%%%%%%%%%%%%

\vspace{-0.7em}
\bibliographystyle{IEEEtran}
\bibliography{references}

\end{document}